\documentclass[10pt,twocolumn,letterpaper]{article}

\usepackage{cvpr}              

\usepackage{graphicx}
\usepackage{amsmath}
\usepackage{amssymb}
\usepackage{booktabs}
\usepackage{placeins} 
\usepackage{float} 
\usepackage{enumitem}
\usepackage{tabularx} 
\usepackage{longtable} 
\usepackage{array} 
\usepackage{ragged2e} 
\usepackage{booktabs} 
\usepackage{placeins}
\usepackage{afterpage} 
\usepackage{caption} 

\usepackage[pagebackref,breaklinks,colorlinks]{hyperref}

\usepackage[capitalize]{cleveref}
\crefname{section}{Sec.}{Secs.}
\Crefname{section}{Section}{Sections}
\Crefname{table}{Table}{Tables}
\crefname{table}{Tab.}{Tabs.}

\def\cvprPaperID{*} 
\def\confName{CVPR}
\def\confYear{2025}

\begin{document}

\title{FlexID: Training-Free Flexible Identity Injection via Intent-Aware Modulation for Text-to-Image Generation}


\author{Guandong Li\\
iFLYTEK\\
\quad (Corresponding Author)\\
\and
Yijun Ding\\
Suning\\
}

\twocolumn[{
\renewcommand\twocolumn[1][]{#1}
\maketitle
}]

\begin{abstract}
Personalized text-to-image generation aims to seamlessly integrate specific identities into arbitrary textual descriptions. However, existing training-free methods typically rely on rigid visual feature injection into the latent space of diffusion models. This blind anchoring mechanism creates a fundamental conflict between identity fidelity and textual context adaptability: strong visual features tend to override the semantic layout of the prompt, leading to rigid expressions, semantic loss, or style inconsistencies when the target text involves complex expressions, poses, or style changes. To overcome this dilemma without introducing expensive fine-tuning costs, we propose FlexID, a novel training-free dual-stream identity modulation framework. We fundamentally decouple identity features into two orthogonal dimensions: a Semantic Identity Projector (SIP) that injects high-level identity priors into the language space through non-destructive residual mapping, establishing macro-level semantic concepts; and a Visual Feature Anchor (VFA) that ensures fine-grained structural fidelity within the latent generation space via cross-attention mechanisms. Crucially, we design a Context-Aware Adaptive Gating (CAG) mechanism. This mechanism simulates human cognitive processes, analyzing the intensity of editing intent in the prompt and the temporal stage of diffusion generation in real-time, dynamically modulating the contribution weights of the two streams. Specifically, when strong editing intent (such as exaggerated expressions or large movements) is detected, CAG automatically relaxes the rigid visual constraints in the latent space and enhances the flexible semantic guidance in the language space, achieving organic synergy between identity preservation and semantic variation. Extensive experiments on the IBench benchmark show that FlexID achieves state-of-the-art balance between identity consistency and textual adherence. Compared to existing methods, FlexID significantly improves editability in complex narrative scenes while maintaining robust identity features, offering an efficient, flexible plug-and-play solution for controllable personalized generation.
\end{abstract}

\section{Introduction}
\label{sec:intro}

Recent advances in text-to-image generation (T2I) have enabled the synthesis of high-quality, high-fidelity images from natural language descriptions. Within this field, personalized identity customization has emerged as one of the most valuable application directions \cite{wang2024instantid, li2025editid, li2025editidv2, guo2024pulid, liu2023facechain}. Its core objective is to preserve the identity features of a specific reference subject while allowing users to freely control the subject's pose, expression, style, and environment through text prompts. This capability shows great potential in digital content creation, immersive storytelling, and film concept design.

To achieve this goal, early fine-tuning-based methods (e.g., \cite{ruiz2023dreambooth, hu2022lora}) excel at identity preservation, but their high training cost and low deployment efficiency limit large-scale application. Therefore, training-free \cite{guo2024pulid, jiang2025infiniteyou} identity injection methods are becoming mainstream. These methods typically leverage visual encoders to extract features from reference images and inject them directly into the latent space of pre-trained diffusion models via attention mechanisms \cite{ye2023ip, mou2024t2i}.

However, existing training-free methods face a fundamental challenge: rigid visual feature injection. Most existing methods treat identity as a set of static, immutable visual patches, forcibly injecting the visual features of the reference image with fixed strength regardless of the context of the text prompt \cite{li2024photomaker, gal2022image}. This blind anchoring mechanism leads to a severe conflict between identity fidelity and text editability. Specifically, when prompts describe high-dynamic expressions (e.g., ``laughing out loud''), large-scale poses (e.g., ``looking back''), or non-photorealistic styles (e.g., ``sketch''), rigid visual features tend to strongly suppress these variations, resulting in generated faces with stiff expressions, semantic misunderstanding, or style inconsistency. Existing solutions are often forced into a zero-sum game between adhering to the reference image and following the prompt, making it difficult to achieve both.

To address this dilemma, we revisit the mechanism of identity features in generative models. We argue that ideal identity injection should not be a single-dimensional physical anchor but hierarchical and fluid. Identity comprises both semantic-level conceptual priors and visual-level structural details. Based on this insight, we propose FlexID, an intent-aware modulated dual-stream identity injection framework. Unlike previous methods that attempt to balance the conflict through global parameter adjustment or data fine-tuning, FlexID introduces a novel paradigm of hierarchical cross-space decoupling, aiming to achieve high-controllable generation without training.

The core architecture of FlexID consists of two orthogonal, synergistic paths. First, we construct a Semantic Identity Projector (SIP) that maps a face image to semantic tokens in the text space and injects them via a non-destructive residual mechanism into the text encoder \cite{liu2025learning,papantoniou2024arc2face}. This provides soft constraints for the identity, enabling the model to understand the macro features of the person at the language level, thereby supporting highly flexible semantic editing \cite{qian2025omni}. Second, we retain a Visual Feature Anchor (VFA) to provide fine-grained hard constraints at the diffusion model's latent space levels, ensuring the physical similarity of facial features.

Crucially, to coordinate these two paths, we design a Context-Aware Adaptive Gating (CAG) mechanism. This module simulates human cognitive processes, analyzing the intensity of editing intent in the prompt (e.g., detecting action or expression verbs) and combining it with the temporal stage of diffusion generation in real-time, dynamically adjusting the weight distribution of the two streams. When strong editing intent is detected, CAG automatically relaxes the rigid visual constraints in the latent space and instead enhances the flexible semantic guidance in the language space, thereby releasing the freedom of semantic generation without destroying identity features.

In summary, the main contributions of this paper are as follows:
\begin{enumerate}
    \item \textbf{FlexID Framework}: The first dual-stream, training-free architecture that decouples identity injection into language-space semantic guidance and latent-space visual anchoring, fundamentally solving the semantic conflict caused by rigid injection.
    \item \textbf{Intent-Aware Dynamic Gating}: A novel inference-time modulation mechanism that adaptively balances semantic and visual contributions based on editing intent and generation timestep, breaking the limitations of traditional static weighting.
    \item \textbf{SOTA Performance}: Extensive experiments on the IBench benchmark show that FlexID, without any fine-tuning, significantly improves editability metrics in complex narrative scenes while maintaining robust identity consistency, achieving a breakthrough in both efficiency and effectiveness.
\end{enumerate}

\section{Related Works}
\label{sec:related_work}

\subsection{Personalized Text-to-Image Generation}
With the rapid development of diffusion models, personalized generation has become a focus of the community \cite{liu2023facechain,li2024commerce,li2024layout,li2023smartbanner,chen2025xverse,mou2025dreamo}. Early mainstream methods primarily relied on test-time fine-tuning. For example, Textual Inversion \cite{gal2022image} represents new concepts by optimizing specific text token embeddings; DreamBooth \cite{ruiz2023dreambooth} implants specific subjects by fine-tuning the entire denoising network; LoRA \cite{hu2022lora} reduces the number of parameters to be fine-tuned via low-rank adaptation. While these methods excel in identity fidelity, they typically require time-consuming training for each identity, and the resulting model files are difficult to reuse, greatly limiting their potential in real-time applications and large-scale deployment. This motivates the research community to explore more efficient training-free solutions, the domain where FlexID resides.

\subsection{Training-Free Identity Preserving Generation}
The core idea of training-free methods is to leverage pre-trained image encoders to extract features from reference images and inject them into diffusion models via specific adapters \cite{xiao2025fastcomposer,he2025uniportrait,li2025dvi,cheng2025umo,wu2025less}. IP-Adapter \cite{ye2023ip} pioneered this direction by introducing image prompts via decoupled cross-attention. Subsequently, InstantID \cite{wang2024instantid} combined the strong control of ControlNet \cite{zhang2023adding} with ID embedding, achieving high-fidelity face generation. PuLID \cite{guo2024pulid} further introduced a contrastive alignment loss and a lightning branch, achieving high-accuracy identity injection with minimal interference to the original model's generative capability.

However, the aforementioned methods mostly employ a single-stream visual feature injection strategy. They tend to treat identity features as static, hard constraints imposed on the latent space. This rigid injection mechanism ensures pixel-level similarity but often overrides semantic information in the text prompt when facing complex textual contexts (e.g., exaggerated expressions, stylized rendering), leading to inflexible results. Unlike these methods, FlexID focuses not only on visual-level anchoring but also introduces semantic-level guidance, solving the editability bottleneck caused by rigid injection through a dynamic gating mechanism.

\subsection{Cross-Modal Semantic Projection and Feature Disentanglement}
To enhance the model's semantic understanding of image content, another line of research focuses on mapping visual features to the semantic space of the text encoder. Works like FaceCLIP \cite{liu2025learning} demonstrate that face images can be projected into text tokens, allowing them to directly participate in the attention computation of the text encoder. PhotoMaker \cite{li2024photomaker} employs stacked ID embeddings to construct identity features.

Although these methods show potential in semantic alignment \cite{mao2024realcustom++,tan2025ominicontrol,tan2025ominicontrol2}, relying solely on semantic-space injection often fails to guarantee fine-grained facial structure reconstruction (i.e., insufficient likeness). The innovation of FlexID is that we do not regard semantic projection as a single solution. Instead, we reframe it as the Semantic Identity Projector (SIP), forming an orthogonal complement with the latent-space Visual Feature Anchor (VFA). More importantly, we introduce a dynamic modulation mechanism, lacking in previous work, enabling the model to adaptively balance the contributions of these two spaces based on semantic intent, thereby achieving a balance between the flexibility of semantic understanding and the fidelity of visual generation.

\section{Method}
\label{sec:method}

\begin{figure*}[t]
  \centering
  \includegraphics[width=\textwidth]{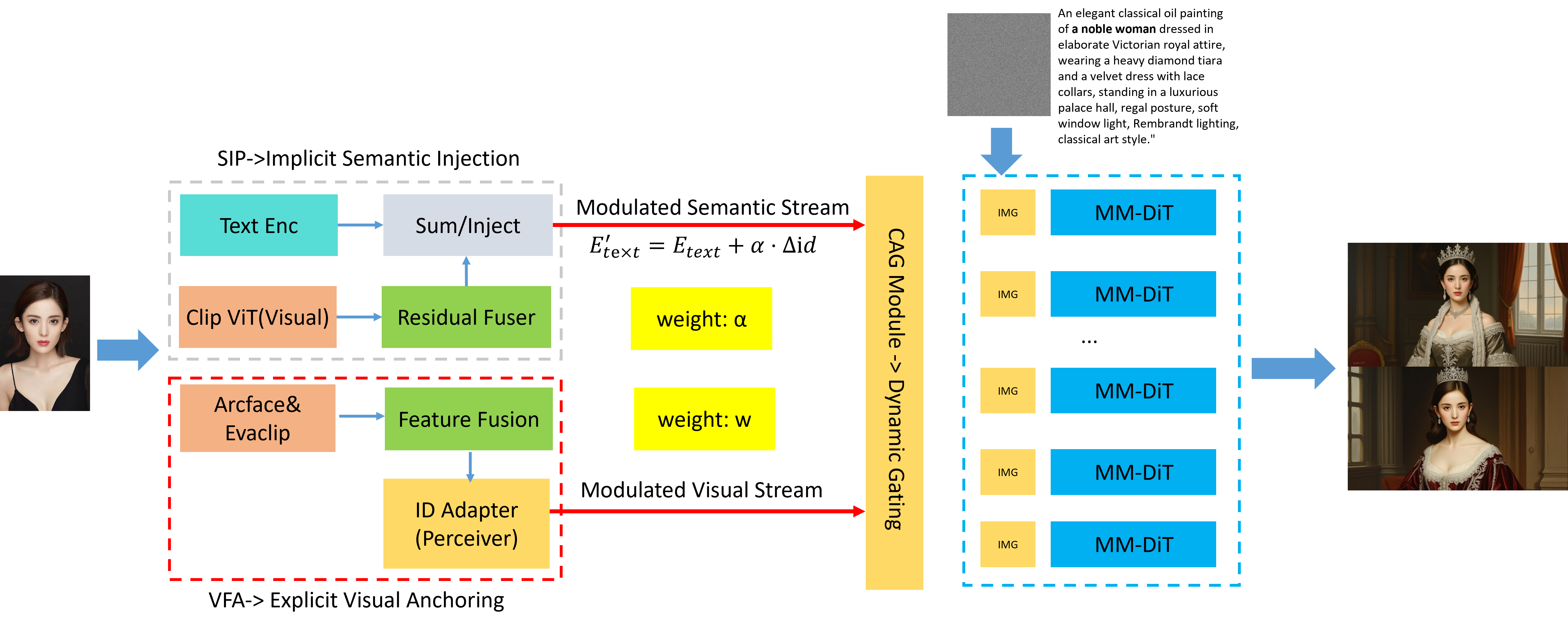} 
  \caption{Overview of the FlexID Framework. Given a reference image, the identity features are disentangled into a Semantic Identity Projector (SIP) and a Visual Feature Anchor (VFA). The SIP provides semantic-level soft guidance to the text encoder, while the VFA provides visual-level structural constraints to the DiT backbone. The Context-Aware Adaptive Gating (CAG) mechanism dynamically balances the contributions of both streams based on the editing intent of the prompt and the current timestep.}
  \label{fig:method}
\end{figure*}

FlexID aims to address the problem of rigid injection and semantic conflict in training-free identity customization. The overall framework is illustrated in \cref{fig:method}. This section first provides an overview of FlexID's dual-stream architecture, then details the two core components: the Semantic Identity Projector (SIP) and the Visual Feature Anchor (VFA), and finally elaborates on the Context-Aware Adaptive Gating (CAG) mechanism that drives their synergy.

\subsection{Overview}
Given a reference face image $I_{ref}$ and a text prompt $P$, our goal is to generate an image $I_{gen}$ that preserves the identity features of $I_{ref}$ while faithfully reflecting the semantic content (including expression, pose, and style) in $P$.

We decouple this process into two orthogonal spaces:
\begin{enumerate}
    \item \textbf{Language Space ($\mathcal{S}_{L}$)}: Responsible for establishing the macro semantic concept of the identity (e.g., ethnicity, gender, general contour).
    \item \textbf{Latent Space ($\mathcal{S}_{Z}$)}: Responsible for anchoring the micro pixel structure of the face (e.g., facial feature shapes, texture details).
\end{enumerate}
The final generation process is controlled by a set of dynamic weights $\{\alpha(t, P), w(t, P)\}$, computed in real-time by the CAG module based on context.

\subsection{Semantic Identity Projector (SIP)}
Traditional identity injection often overlooks the role of the text encoder. To endow the model with semantic understanding of identity, we construct the SIP module.

SIP utilizes a pre-trained visual encoder (Vision Transformer) to extract features $f_{vis}$ from the reference image $I_{ref}$. To achieve cross-modal alignment, we introduce a lightweight Transformer Fuser as the projection head $\Phi_{proj}$, mapping visual features to the text feature space:
\[
\Delta_{id} = \Phi_{proj}(f_{vis}, E_{text}),
\]
where $E_{text}$ is the original embedding of the text prompt $P$.

\textbf{Non-Destructive Residual Mapping}:
To avoid overwriting the non-identity semantics (e.g., adjectives, action verbs) in $P$, we do not directly replace text tokens but adopt a residual injection strategy:
\[
E'_{text} = E_{text} + \alpha \cdot \Delta_{id},
\]
where $\alpha$ is the semantic guidance coefficient. As $\alpha$ is typically set to a small value ($\alpha \in [0.05, 0.15]$), this design constitutes a soft identity constraint, allowing the model to make slight semantic bias of the generative manifold towards the identity direction while preserving the original syntactic structure.

\subsection{Visual Feature Anchor (VFA)}
While SIP provides macro-level guidance at the semantic level, to ensure pixel-level consistency in generated facial details (e.g., facial feature proportions, bone structure), we need a strong visual anchoring mechanism. The VFA module aims to provide high-fidelity hard constraints in the latent space.

\textbf{Dual-Granularity Feature Extraction and Disentanglement}:
Referencing advanced feature extraction strategies, VFA adopts a collaborative global-local feature extraction scheme to capture complete identity information:
\begin{enumerate}
    \item \textbf{Global Features}: We utilize the ArcFace encoder to extract deep semantic features of the face. These features encode the overall facial structure and identity ID information, characterized by high coupling and stability, and are responsible for maintaining the ID consistency of the generated image.
    \item \textbf{Local Features}: We further utilize EvaCLIP to extract fine-grained local visual features. By filtering specific Identity-Aware layers, these features capture facial texture, lighting, shadow, and micro-expression details, carrying more visual editability.
\end{enumerate}
After fusion, the two sets of features form the complete identity visual embedding $V_{id}$.

\textbf{Injection Mechanism Based on Contrastive Alignment}:
To inject these features into the generative process of Flux DiT with minimal distortion, VFA draws inspiration from the design of the Lightning T2I Branch. This mechanism, by introducing a contrastive alignment loss during the training phase, minimizes interference from ID insertion on the original model's generative behavior while enhancing fidelity using ID loss.

During inference, VFA injects $V_{id}$ into specific layers of DiT (typically ID-Sensitive Blocks) via decoupled cross-attention. In the $l$-th DiT layer, the injection process is formulated as:
\[
\begin{aligned}
\text{Attention}(Q, K, V) &= \text{Softmax}\!\left(\frac{Q K^T}{\sqrt{d}}\right)V \\
&\quad + w \cdot \text{Attention}(Q, K_{id}, V_{id}),
\end{aligned}
\]
where $K_{id}, V_{id}$ are the key-value pairs projected from $V_{id}$.

\textbf{From Rigid to Flexible}:
In traditional training-free methods, the injection weight $w$ is typically set to a high fixed value (e.g., $w=1.0$). This rigid injection, while ensuring high ID reconstruction accuracy, can easily disrupt the balance of the noise distribution when facing complex text contexts, leading to loss of lighting or semantic override.
The core innovation of VFA in FlexID lies in transforming this carefully designed injection module from absolute dominance to controlled anchoring. The weight $w$ is no longer a constant but becomes a dynamic variable $w(t, P)$ modulated by the CAG module, thereby preserving high-fidelity visual features while leaving room for semantic variation.

\subsection{Context-Aware Adaptive Gating (CAG)}
This is the core innovation of FlexID. CAG aims to break the static balance between SIP and VFA, achieving dynamic modulation at inference time. CAG consists of two sub-modules: semantic intent parsing and temporal complementary scheduling.

\subsubsection{Semantic-Driven Intent Adaptation}
We observe that the conflict intensity between identity consistency and editability freedom depends on the editing intent of the prompt $P$. Rigid visual anchoring hinders generation when $P$ contains descriptions of strong expressions or actions.

CAG first extracts semantic attributes from $P$ via a keyword parser, defining an intent indicator function $\mathbb{I}_{intent}(P)$:
\[
\mathbb{I}_{intent}(P) =
\begin{cases}
1, & \text{if } P \cap \mathcal{D}_{edit} \neq \emptyset \text{ (High-Edit Intent)} \\
0, & \text{otherwise (Low-Edit Intent)}
\end{cases}
\]
where $\mathcal{D}_{edit}$ is a semantic dictionary containing expressions (e.g., \textit{smile, cry}), poses (e.g., \textit{turn, run}), and stylistic descriptions.

Based on the intent detection result, we define dynamic adjustment factors:
\begin{align*}
\gamma_{sem} &= 1 + \lambda_{up} \cdot \mathbb{I}_{intent}(P), \\
\gamma_{vis} &= 1 - \lambda_{down} \cdot \mathbb{I}_{intent}(P).
\end{align*}
When high editing intent is detected, CAG automatically increases the semantic stream weight ($\gamma_{sem} > 1$) and suppresses the visual stream weight ($\gamma_{vis} < 1$). This allows the model to switch from pixel-copying mode to semantic-imagination mode.

\subsubsection{Diffusion-Stage Complementary Scheduling}
Inspired by the dynamics of the diffusion generation process—which follows a pattern from coarse semantic layout to fine texture refinement—we introduce a complementary scheduling strategy over the generation timestep $t$.

We design two temporal functions:
\begin{itemize}
    \item Semantic decay function $\mathcal{T}_{sem}(t) \propto t$: Assigns higher weight to SIP in the early generation stage ($t \to T$) to establish composition and macro attributes.
    \item Visual growth function $\mathcal{T}_{vis}(t) \propto (1-t)$: Assigns higher weight to VFA in the late generation stage ($t \to 0$) to refine high-frequency facial details.
\end{itemize}

\subsubsection{Final Dynamic Weight Formula}
Integrating the two dimensions above, the final injection weights for FlexID at any timestep $t$ and prompt $P$ are:
\begin{align}
\alpha_{final}(t, P) &= \alpha_{base} \cdot \gamma_{sem}(P) \cdot \mathcal{T}_{sem}(t), \\
w_{final}(t, P) &= w_{base} \cdot \gamma_{vis}(P) \cdot \mathcal{T}_{vis}(t).
\end{align}
Through this mechanism, FlexID achieves context-aware and time-dependent flexible identity injection, solving the semantic conflict problem caused by rigid injection without any fine-tuning.

\section{Experiments}
\label{sec:experiments}

\subsection{Experimental Settings}
\textbf{Implementation Details}. Our method is implemented based on the Flux.1-dev model. For the Semantic Identity Projector (SIP), we utilize the Patch Tokens of CLIP ViT-L/14 \cite{radford2021learning} for feature extraction and project them into the text space via the Residual Fuser from FaceCLIP. For the Visual Feature Anchor (VFA), we reuse the architecture of PuLID, employing ArcFace and EvaCLIP \cite{sun2023eva} to extract high-fidelity visual features.
During inference, the default sampling steps are set to $T=25$, Guidance Scale to 4.0, and the sampler is Euler. To allow dynamic modulation space for the Context-Aware Adaptive Gating (CAG), we set the base injection weight for VFA to $1.0$ and the base residual weight for SIP to $0.1$. All experiments are conducted on a single NVIDIA A100 (80GB) GPU, and the evaluation framework is IBench.

\subsection{Qualitative Comparison}
\begin{figure*}[t]
    \centering
    \includegraphics[width=\textwidth]{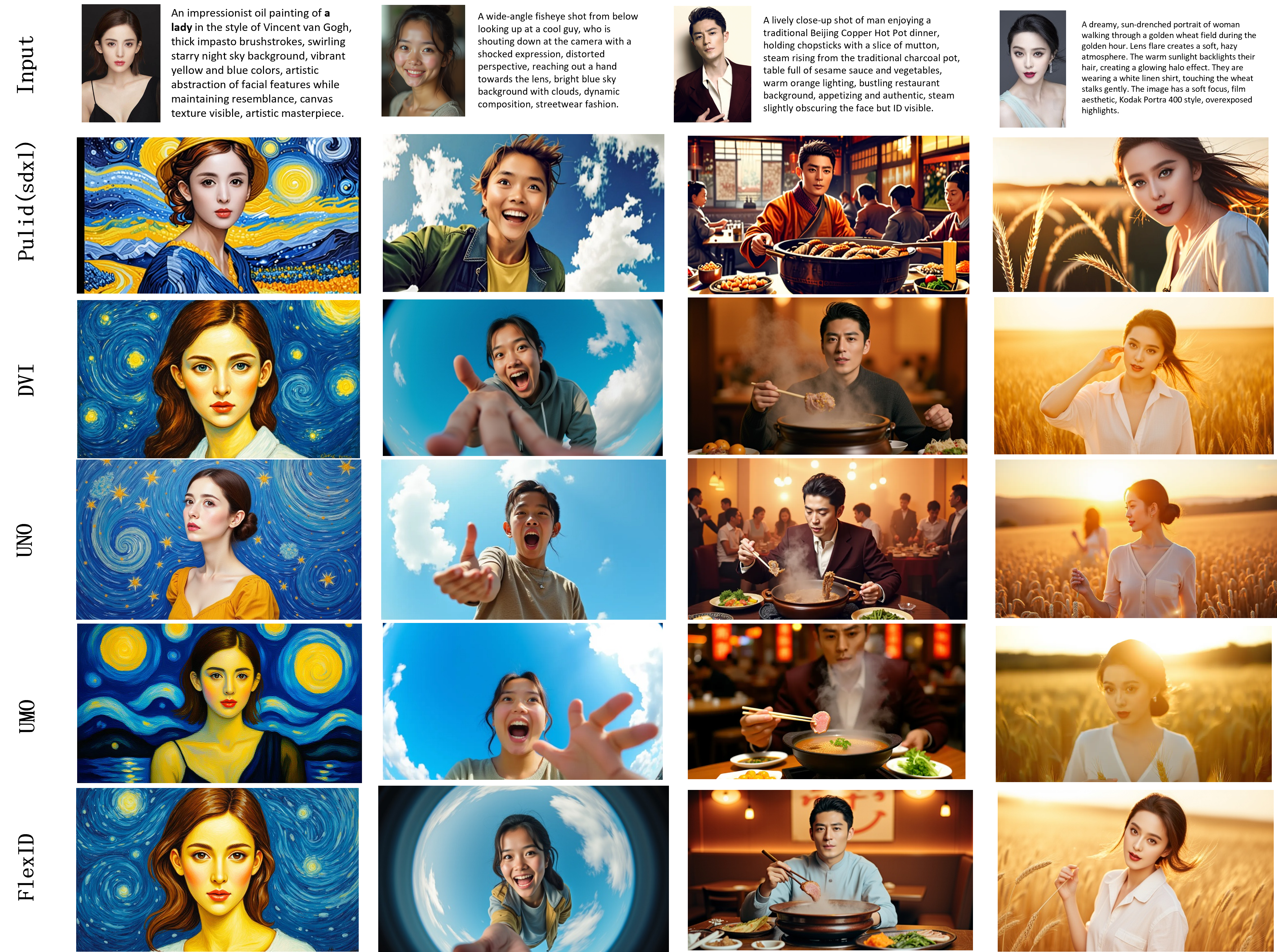}
    \caption{Qualitative Comparison. FlexID achieves superior visual atmosphere integration and identity-consistent editability in complex scenes (e.g., stylized oil painting, fisheye lens, atmospheric dining, backlit portrait) compared to other tuning-free methods.}
    \label{fig:qualitative}
\end{figure*}

We adopt the SDXL \cite{podell2023sdxl} version of PuLID, as well as DVI, UNO, and UMO as the comparison model group. The base model for PuLID (SDXL) is SDXL\_base\_1.0, while the base models for all other models (including FlexID) are Flux.1 dev. As shown in \cref{fig:qualitative}, we input long text prompts containing strong stylistic descriptions, extreme geometric perspectives, and complex lighting interactions. While maintaining character consistency, compared to PuLID (SDXL) and DVI, FlexID achieves superior Visual Atmosphere Integration; compared to UNO and UMO, FlexID not only avoids severe ID loss but also generates images with more natural texture, spatial structure, and strict adherence to the environment and style instructions in the prompt. \Cref{fig:qualitative} showcases four highly challenging generation scenarios: Impressionist oil painting, wide-angle fisheye exclamation, street hotpot dinner, and backlit wheat field portrait. In the first column (Impressionist oil painting), the prompt explicitly requests ``thick impasto brushstrokes... swirling starry night sky background''. The faces generated by PuLID (SDXL) and DVI are overly clean and smooth, presenting a jarring feeling of a modern digital photo forcibly pasted onto a painted background, completely disconnected from the brushstroke texture of the background. While UNO generates a strong artistic style, the character's ID fundamentally drifts. In contrast, FlexID successfully ``injects'' the thick, textured brushstrokes of the Van Gogh style into the facial skin; the facial features are recomposed with colorful strokes, preserving the bone structure features of the reference image while perfectly fitting the overall painting style of a ``masterpiece''. In the second column (fisheye exclamation), when faced with the instruction ``Wide-angle fisheye shot... distorted perspective'' demanding extreme geometric deformation, DVI and UMO exhibit significant rigidity, maintaining relatively standard perspective proportions in the face, failing to capture the exaggerated visual impact of ``fisheye''. In contrast, FlexID accurately recognizes the intent of image distortion, allowing the face to produce natural barrel distortion with the lens while maintaining ID recognizability, and presenting a highly expressive expression.

\subsection{Quantitative Evaluation}
We further validate the above visual observations using the metrics from the ChineseID + Editable Long Prompts benchmark test, with specific data shown in \cref{tab:results}. FlexID demonstrates excellent comprehensive performance across multiple key metrics, successfully balancing identity fidelity, visual aesthetics, and text adherence under the zero-shot setting.

In terms of Aesthetic Quality \& Visual Atmosphere, FlexID achieves the highest score on the Aesthetic (0.695) metric among all compared models (all based on Flux architecture), outperforming Dreamo (0.678) and PuLID (Flux) (0.681). Meanwhile, on Image Quality measuring overall image quality, FlexID also reaches 0.499, on par with the strong-performing Dreamo (0.510) and significantly higher than PuLID (Flux) (0.430). This strongly proves the effectiveness of the CAG dynamic gating mechanism—by timely relaxing visual constraints during generation, FlexID successfully eliminates the ``stiffness'' common in traditional ID injection methods, making the generated images more aesthetically pleasing in terms of lighting, texture, and artistic composition, rather than mere feature collages.

Regarding the trade-off between Identity Preservation vs. Editability, FlexID demonstrates a highly competitive balancing strategy. On the Facesim metric measuring identity consistency, FlexID reaches 0.671. This is a critical watershed: Dreamo, also based on Flux architecture, suffers a drastic drop to 0.398 in Facesim, indicating severe identity distortion in its generated images. In contrast, while PuLID (Flux) achieves a high FaceSim of 0.735, its Landmarkdiff is only 0.066, hinting at severe ``anatomical locking''. FlexID maintains a higher Landmarkdiff (0.073) and more flexible pose while its ID fidelity is almost double that of Dreamo. This reflects that FlexID occupies a ``golden balance point'': it does not lose ID for the sake of aesthetics like Dreamo, nor does it sacrifice facial dynamism for ID preservation like PuLID.

In terms of Visual \& Textual Consistency, FlexID demonstrates a deep understanding of long-text contexts. On the ClipT metric measuring textual adherence, FlexID reaches 0.261, significantly outperforming PuLID (Flux) (0.247) and approaching Dreamo (0.266), which trades ID loss for better text alignment. This proves the effectiveness of the SIP module, which helps the model not drown out non-identity semantic details in the prompt while injecting identity. Meanwhile, on the ClipI metric measuring global visual style consistency, FlexID maintains a high value of 0.769, indicating that it does not lose the visual essence of the reference image due to enhanced semantic understanding.

In summary, FlexID does not simply pursue extreme values in a single metric but finds the optimal Pareto Frontier between identity fidelity (Facesim), visual aesthetics (Aesthetic), and text adherence (ClipT). Especially in high-complexity narrative scenes, this balancing ability enables the images generated by FlexID to not only ``look like the person'' but also possess narrative tension that ``matches the description''.

\begin{table*}[t]
    \centering
    \caption{Evaluation metric results from IBench on ChineseID with editable long prompts.}
    \label{tab:results}
    \begin{tabular}{lcccccc}
        \toprule
        Model & Aesthetic & Image quality & Landmarkdiff & Facesim & ClipI & ClipT \\
        \midrule
        PuLID (SDXL) & 0.675 & 0.502 & 0.099645292 & 0.399 & 0.768 & 0.248 \\
        PuLID (Flux) & 0.6809378 & 0.430490792 & 0.0657981 & 0.735393405 & 0.7498004 & 0.246589 \\
        Dreamo & 0.67795705 & 0.509961 & 0.11130033 & 0.397947 & 0.8045682 & 0.265729 \\
        UNO & 0.6753485 & 0.46532 & 0.11643834 & 0.1050183 & 0.797489 & 0.266868 \\
        UMO & 0.668947458 & 0.469073 & 0.1128649 & 0.397358 & 0.7480032 & 0.258867 \\
        FlexID (Ours) & \textbf{0.695377} & 0.498637 & 0.073359 & 0.6707494 & 0.76894116 & 0.26055799 \\
        \bottomrule
    \end{tabular}
\end{table*}

\subsection{Ablation Study}
To deeply investigate the role of each component in FlexID, we conducted detailed ablation studies.

\subsubsection{Effectiveness of Semantic Identity Projector (SIP)}
\begin{figure}[t]
\centering
\includegraphics[width=\linewidth]{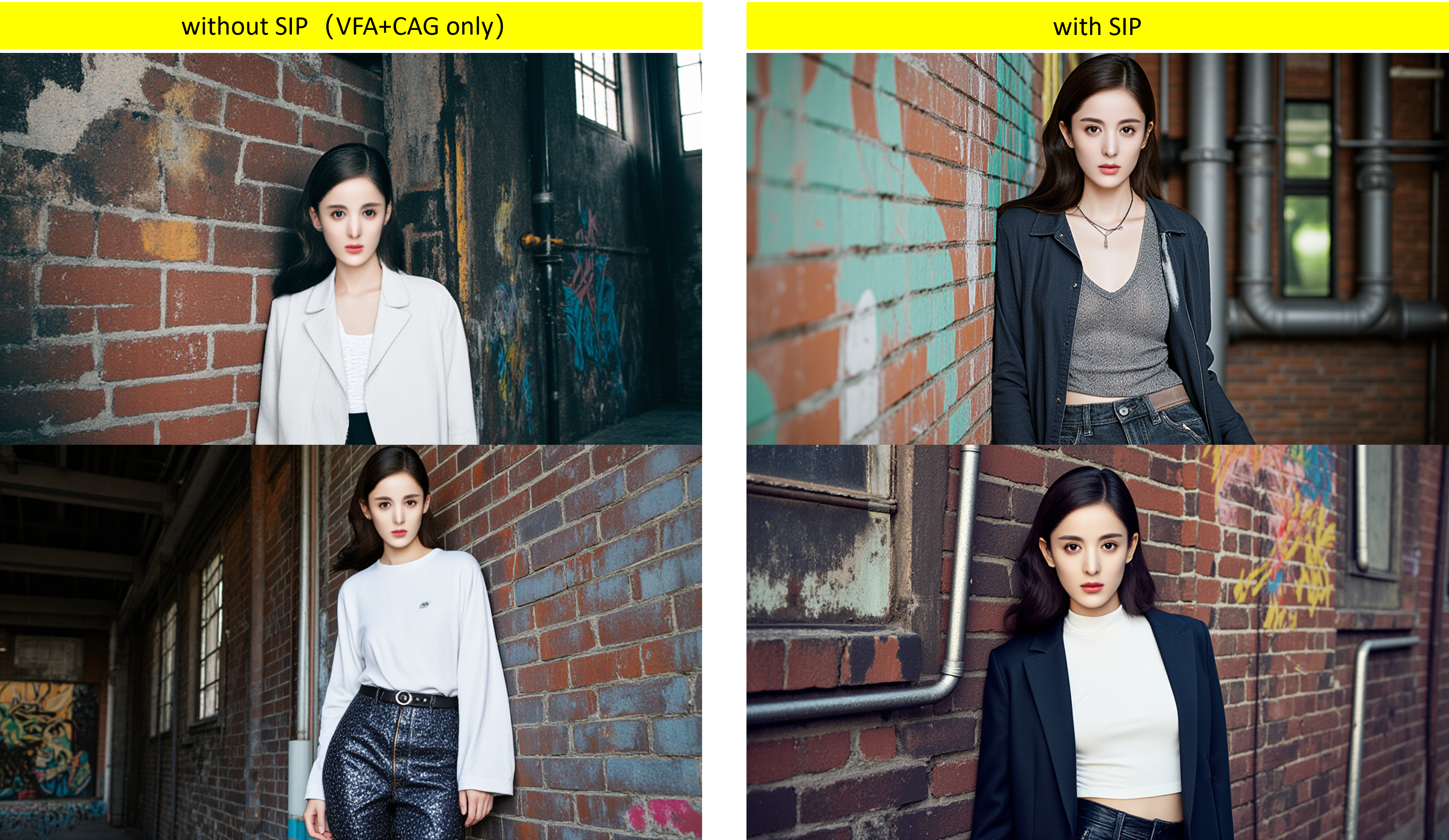}
\caption{Effectiveness of the Semantic Identity Projector (SIP). Left: Without SIP, the identity appears isolated from the environment. Right: With SIP, the identity naturally integrates with the environmental atmosphere and style.}
\label{fig:ablation_sip}
\end{figure}

To verify the core role of the SIP module in establishing macro semantic concepts, we conducted an ablation experiment by removing the SIP module and observing the generation results with only the CAG-controlled VFA retained. As shown in \cref{fig:ablation_sip}, when processing a prompt describing an industrial fashion shoot, the result without SIP (left image) shows a clear Contextual Isolation phenomenon: although VFA strongly anchors the facial features, the lighting on the person's face is flat and lacks depth, and the clothing is just a basic white shirt, appearing like a ``sticker'' forcibly pasted onto a textured red brick background, severely disconnected from the environmental atmosphere. In contrast, with the SIP module introduced (right image), the identity features are mapped into fluid semantic concepts, successfully stimulating the model's deeper understanding of ``fashion'' and ``atmosphere'' in the prompt. The generated image not only features darker, more textured clothing but also naturally integrates facial lighting into the low-key industrial environment. This comparison strongly proves that SIP acts as a semantic bridge between visual features and textual descriptions, ensuring the person organically adapts to complex environmental lighting and stylistic narratives while preserving ID.

\subsubsection{Core Role of Intent-Aware Gating (CAG)}
\begin{figure}[t]
\centering
\includegraphics[width=\linewidth]{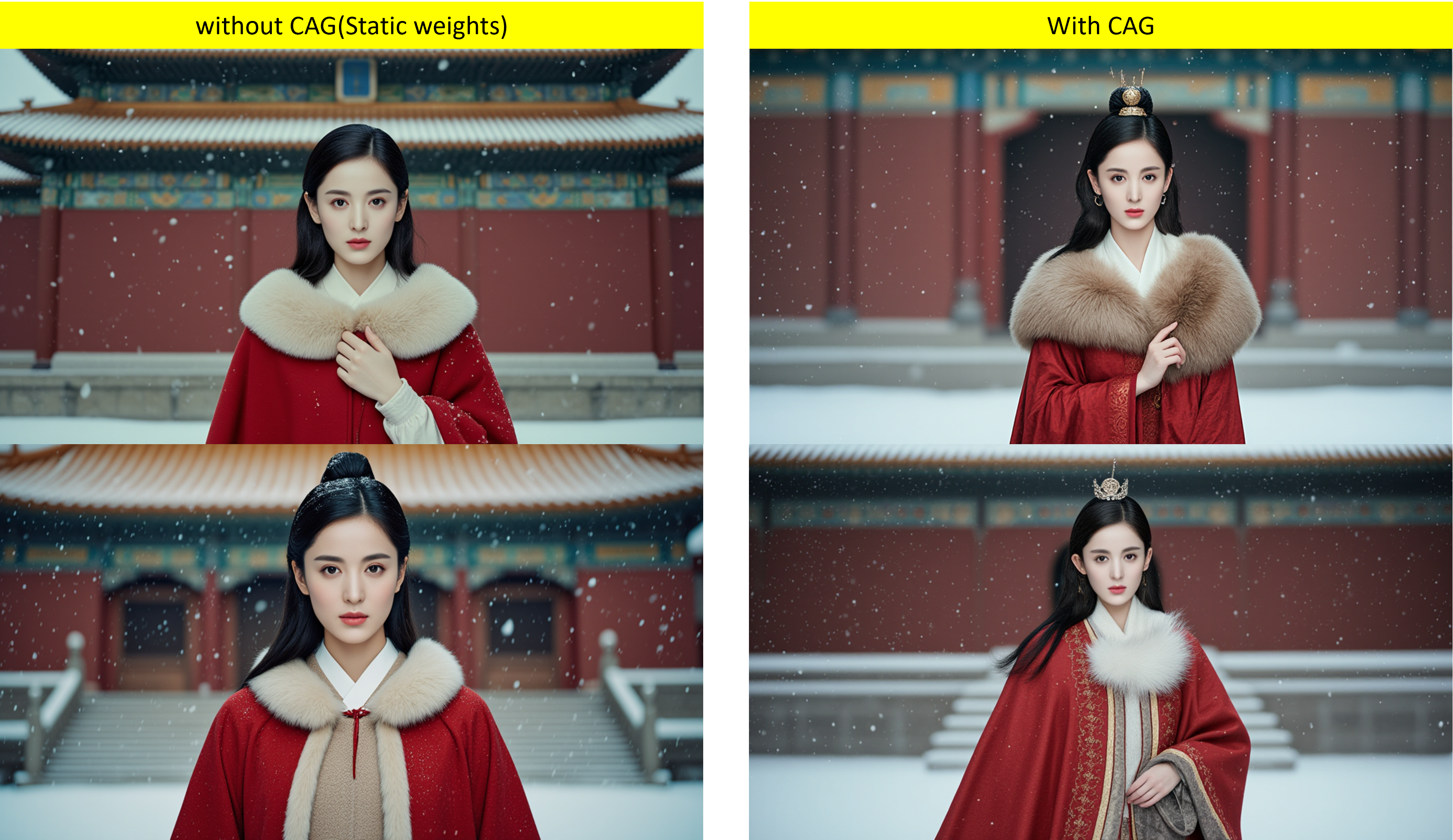}
\caption{Core Role of the Intent-Aware Gating (CAG) Mechanism. Left: Static weighting leads to rigid poses and loss of aesthetic details. Right: CAG enables flexible pose and enriches details while preserving ID.}
\label{fig:ablation_cag}
\end{figure}

We further compared the effects of static weighting strategies versus the CAG dynamic gating mechanism. As shown in \cref{fig:ablation_cag}, when generating the highly aesthetically demanding scene ``ancient costume portrait in the Forbidden City with first snow'', the result using fixed weights (Without CAG, left image) falls into the typical rigid anchoring trap: due to overly strong visual constraints in the latent space, the model is forced to prioritize preserving the physical structure of the face, resulting in a stiff, ``ID-photo-style'' frontal standing pose. To accommodate ID stability, the model sacrifices the clothing details described in the prompt, making the red cloak appear plain. Conversely, the CAG mechanism of FlexID (right image) successfully breaks this stalemate. CAG keenly captures the aesthetic intent in the prompt and dynamically adjusts the dual-stream weights during generation. This not only makes the character's pose natural and relaxed (presenting a cinematic side-profile composition) but, more importantly, releases the generative freedom of the semantic space—allowing the exquisite golden headpiece, intricate embroidery on the cloak, and more realistic fur texture to be perfectly presented. This mechanism is like a photographer who masters the art of ``capture and release'', maximizing the organic growth of environmental atmosphere and costume details while ensuring high identity recognizability, achieving dynamic unity between identity features and complex narrative aesthetics.

\section{Conclusion}
This paper presents FlexID, a training-free identity customization framework for high-complexity narrative scenes. Addressing the fundamental conflict between Rigid Visual Injection and Semantic Context Adaptation prevalent in existing methods, we innovatively introduce a hierarchical cross-space identity modulation paradigm.

By decoupling identity features into the Semantic Identity Projector (SIP) in the language space and the Visual Feature Anchor (VFA) in the latent space, and employing a Context-Aware Adaptive Gating (CAG) mechanism to drive their dynamic synergy, FlexID successfully breaks the shackles of ``anatomical locking'' and ``semantic islands'' inherent in traditional methods. Experimental results show that FlexID achieves state-of-the-art balance between identity fidelity and text editability on the IBench benchmark with only a single forward pass. Compared to methods requiring expensive data fine-tuning, FlexID proves that through sophisticated inference-time architectural design, the deep semantic potential of generative models can be fully activated in a zero-shot setting.

We believe that the paradigm shift from rigid anchoring to dynamic modulation advocated by FlexID not only provides an efficient, plug-and-play solution for personalized text-to-image generation but also offers inspiring theoretical reference for future controllable video generation and multimodal editing tasks. In future work, we will further explore how to extend this dynamic gating mechanism to more challenging scenarios such as multi-subject interaction and fine-grained attribute disentanglement.

\section*{Declarations}
\subsection*{Funding and/or Conflicts of interests/Competing interests}
The authors declare that they have no known competing financial interests or personal relationships that could have appeared to influence the work reported in this paper. The authors did not receive support from any organization for the submitted work.

\subsection*{Ethical and informed consent for data used}
This article does not contain any studies with human participants or animals performed by any of the authors. The research utilizes publicly available datasets (IBench) and pre-trained models. Therefore, ethical approval and informed consent were not required.

\subsection*{Author Contribution}
Guandong Li proposed the original idea, designed the experiments, and wrote the manuscript. Yijun Ding was responsible for implementing specific experimental modules and conducting part of the experimental analysis.

\subsection*{Data Availability}
The datasets generated during and/or analysed during the current study are available in the \textbf{IBench} repository (https://github.com/typemovie/IBench). The pre-trained base models (Flux.1, ArcFace, CLIP, EvaCLIP) used in this study are publicly available from their respective repositories.

{\small
\bibliographystyle{ieee_fullname}
\bibliography{egbib}
}

\end{document}